\documentclass[sigconf]{acmart}
\AtBeginDocument{%
  }

\usepackage{algorithm}
\usepackage{algpseudocode}
\usepackage{multirow} 
\usepackage{float}
\usepackage{booktabs}
\usepackage{balance}

\copyrightyear{2026}
\acmYear{2026}
\setcopyright{cc}
\setcctype{by}
\acmConference[KDD 2026]{Proceedings of the 32nd ACM SIGKDD Conference on 
Knowledge Discovery and Data Mining V.2}{August 9--13, 2026}{Jeju Island, 
Republic of Korea.}
\acmBooktitle{Proceedings of the 32nd ACM SIGKDD Conference on Knowledge 
Discovery and Data Mining V.2 (KDD 2026), August 9--13, 2026, Jeju Island, 
Republic of Korea}
\acmISBN{979-8-4007-2259-2/2026/08}
\acmDOI{10.1145/3770855.3818316}

\begin{document}

\title{AIGP: An LLM-Based Framework for Long-Term Value Alignment in E-Commerce Pricing}


\author{Chennan Ma}
\authornote{Both authors contributed equally to this work.}
\email{machennan.mcn@alibaba-inc.com}
\affiliation{%
  \institution{Taobao \& Tmall Group of Alibaba}
  \city{Hangzhou}
  \state{Zhejiang}
  \country{China}
}

\author{Yanning Zhang}
\authornotemark[1]
\email{zhangyanning.zyn@taobao.com}
\affiliation{%
  \institution{Taobao \& Tmall Group of Alibaba}
  \city{Hangzhou}
  \state{Zhejiang}
  \country{China}
}

\author{Siqi Hong}
\email{hongsiqi.hsq@taobao.com}
\affiliation{%
  \institution{Taobao \& Tmall Group of Alibaba}
  \city{Hangzhou}
  \state{Zhejiang}
  \country{China}
}

\author{Xiuchong Wang}
\authornote{Corresponding author.}
\email{xiuchong.wxc@alibaba-inc.com}
\affiliation{%
  \institution{Taobao \& Tmall Group of Alibaba}
  \city{Hangzhou}
  \state{Zhejiang}
  \country{China}
}

\author{Fei Xiao}
\email{guren.xf@alibaba-inc.com}
\affiliation{%
  \institution{Taobao \& Tmall Group of Alibaba}
  \city{Hangzhou}
  \state{Zhejiang}
  \country{China}
}

\author{Keping Yang}
\email{shaoyao@taobao.com}
\affiliation{%
  \institution{Taobao \& Tmall Group of Alibaba}
  \city{Hangzhou}
  \state{Zhejiang}
  \country{China}
}


\renewcommand{\shortauthors}{Chennan Ma et al.}

\begin{abstract}
  Traditional dynamic pricing models in large-scale e-commerce suffer from limited interpretability, poor utilization of unstructured information, and misalignment with long-term business objectives such as cumulative Gross Merchandise Value (GMV), Return on Investment (ROI) and milestone achievement. We propose \textbf{AIGP}, a novel framework that leverages a Large Language Model (LLM) prompted with domain knowledge, structured data and textual context to make interpretable, knowledge-aware pricing decisions. For efficient deployment while maintaining high-quality outputs, we employ supervised fine-tuning for knowledge distillation. Central to AIGP is the Long-Term Value Estimator (LTVE), trained via offline reinforcement learning on historical data, which serves as a reward model to score candidate pricing actions and select preference pairs for Direct Preference Optimization (DPO), thereby aligning the pricing policy with long-term business objectives. Extensive offline evaluations and large-scale online A/B tests on Tao Factory demonstrate that AIGP achieves significant improvements: \textbf{+13.21\%} in GMV, \textbf{+7.59\%} in ROI, and \textbf{+8.20\%} in milestone achievement rate over 14 days compared to the production baseline, while simultaneously providing interpretable and transparent pricing rationales.
\end{abstract}




\maketitle

\section{Introduction}
Dynamic pricing is a critical driver of market efficiency across many industries, with established applications in airlines, hotels, and ride-hailing~\cite{talluri2004revenue,mcgill1999revenue,gallego1994optimal,ferreira2016analytics}. In large-scale e-commerce platforms that manage extensive Stock Keeping Units (SKUs) and high-frequency daily interactions~\cite{chen2016empirical,Chen2016AnEA}, effective dynamic pricing mechanisms are essential for efficiently responding to rapidly changing market conditions and consumer preferences.

Traditional solutions fall into two categories: rule-based systems built upon handcrafted heuristics, and data-driven models leveraging price–sales elasticity estimation and mathematical optimization~\cite{gallego1994optimal,ferreira2016analytics}. However, both approaches suffer from limited decision transparency, poor utilization of unstructured information (e.g., product descriptions, user reviews), and overemphasis on immediate gains. Aggressive discounting to boost short-term sales can erode margins and constrain future operations, making it difficult to optimize long-horizon objectives such as cumulative Gross Merchandise Value (GMV), Return on Investment (ROI), and long-term milestone achievement.

Reinforcement learning (RL) has recently been applied to dynamic pricing, allowing direct optimization for long-term business objectives through value function modeling, which estimates the discounted cumulative return of future outcomes~\cite{ddpg2019dynamic,sac2024dynamic,maddpg2025dynamic, railpricing2025marl}. Yet RL inherits issues with interpretability and limited use of unstructured data, and introduces further challenges like distributional shift~\cite{kumar2020cql,levine2020offline,fujimoto2019off} and reward sparsity~\cite{andrychowicz2017hindsight,pathak2017curiosity,burda2018exploration}, especially for cold-start or long-tail products~\cite{yin2012challenging,ji2021reinforcement,wang2017biucb}.

Recent advances in large language models (LLMs) offer new possibilities for transparent and knowledge-rich decision making~\cite{touvron2023llama,wang2022self,bubeck2023sparks,casper2023open}. 
LLMs’ chain-of-thought reasoning provides auditable explanations for pricing decisions, while their ability to process structured and textual inputs enables richer context integration. Moreover, their reasoning over text and analogical knowledge can improve robustness for cold-start and out-of-distribution products. However, generic LLM agents lack platform-specific operational knowledge and supervision from expert demonstrations or long-horizon outcome feedback, leading to suboptimal, short-sighted actions that fail to optimize long-term business objectives.

We propose \textbf{AIGP} (\textbf{A}rtificial \textbf{I}ntelligence \textbf{G}enerated \textbf{P}ricing), a framework that integrates LLM-based reasoning with long-term business value alignment for dynamic pricing. AIGP employs a carefully designed prompt incorporating chain-of-thought~\cite{wei2022cot} (CoT) reasoning, structured signals, domain knowledge, and textual context to generate interpretable, actionable pricing decisions. For efficient deployment, we use supervised fine-tuning~\cite{ouyang2022instructgpt,taori2023alpaca} (SFT) with teacher-student distillation~\cite{hinton2015distilling,xu2024distilling}, enabling compact models to produce high-quality reasoning while satisfying business constraints. Central to AIGP is the \textbf{Long-Term Value Estimator (LTVE)}, trained via offline RL on historical trajectories to model the long-horizon business impact of pricing decisions beyond immediate sales. By scoring candidate pricing actions, LTVE automates preference pair selection for Direct Preference Optimization~\cite{rafailov2024dpo} (DPO), aligning the pricing mechanism with long-term objectives while preserving interpretability.

We deploy AIGP on Tao Factory and validate its performance through comprehensive offline evaluations and large-scale online A/B tests. AIGP consistently outperforms traditional and RL-based baselines, achieving \textbf{+13.21\%} in GMV, \textbf{+7.59\%} in ROI, and \textbf{+8.20\%} in milestone achievement rate over 14 days, while providing interpretable pricing rationales.

\textbf{Our main contributions:}
(1) We propose \textbf{AIGP}, an LLM-based dynamic pricing framework integrating long-horizon business reward modeling for transparent, business-aligned decisions.
(2) We develop a preference alignment pipeline, leveraging an offline RL-trained LTVE to automate preference pair generation for DPO.
(3) We demonstrate robust, scalable deployment via distillation-based SFT, enabling compact models for large-scale production use.

\section{Related Work}
\subsection{Traditional and RL-based Pricing Methods}
Early dynamic pricing methods in e-commerce largely rely on rule-based heuristics and demand estimation models using structured historical data~\cite{talluri2004revenue,mcgill1999revenue,gallego1994optimal,ferreira2016analytics}. For example,~\cite{talluri2004revenue} formalizes threshold-based revenue management, while~\cite{gallego1994optimal} employs demand curves for inventory pricing. These approaches typically focus on short-term profit maximization and struggle with cold-start or long-tail scenarios. More critically, they cannot leverage rich unstructured information such as product descriptions and user reviews, which contain valuable signals for pricing decisions~\cite{phillips2021pricing,keskin2014dynamic,besbes2009dynamic}.

Reinforcement learning (RL)~\cite{mnih2015dqn,lillicrap2016ddpg,haarnoja2018sac,riskq,chen2021decisiontransformer} has been increasingly applied to dynamic pricing by modeling the problem as a Markov decision process and optimizing policies for long-term business objectives via value functions~\cite{ddpg2019dynamic,sac2024dynamic,maddpg2025dynamic,railpricing2025marl}. Representative approaches include DDPG~\cite{ddpg2019dynamic}, SAC~\cite{sac2024dynamic}, and multi-agent approaches~\cite{maddpg2025dynamic,railpricing2025marl} such as MADDPG~\cite{maddpg}, MADQN~\cite{madqn}, and QMIX~\cite{qmix}. However, RL methods inherit limitations from traditional approaches in handling unstructured data and introduce additional challenges: lack of interpretability—critical for production deployment—and instability under distributional shifts in offline settings~\cite{levine2020offline,fujimoto2019off,kumar2020cql}.

\subsection{LLMs for Decision Making and Preference Alignment}
Large language models exhibit strong reasoning capabilities through chain-of-thought prompting~\cite{wei2022cot}, tool augmentation~\cite{schick2023toolformer}, and program synthesis~\cite{nye2021candprograms}, with successful applications in planning and control tasks~\cite{huang2022language,yao2023react,ma2024large,wang2023voyager}. LLMs naturally handle both structured and unstructured inputs while providing interpretable reasoning traces, which are key advantages over black-box approaches. However, their application to dynamic pricing in e-commerce remains unexplored, and generic LLMs lack domain-specific operational knowledge and supervision from long-horizon outcome feedback, often leading to suboptimal, short-sighted decisions.

Aligning LLM outputs with task-specific objectives is critical for production deployment. Supervised fine-tuning (SFT)~\cite{ouyang2022instructgpt,taori2023alpaca} improves instruction-following, it does not optimize for long-term outcomes. Direct Preference Optimization (DPO)~\cite{rafailov2024dpo} enables preference alignment, yet scaling high-quality preference data for long-term value remains challenging~\cite{ziegler2019humanprefs,stiennon2020summarize}. We address this by integrating offline RL-based value estimation with LLM reasoning for business-aligned dynamic pricing. To our knowledge, this is the first such framework in e-commerce.

\section{Methodology}
\begin{figure*}[t]
\centering
\includegraphics[width=0.85\linewidth]{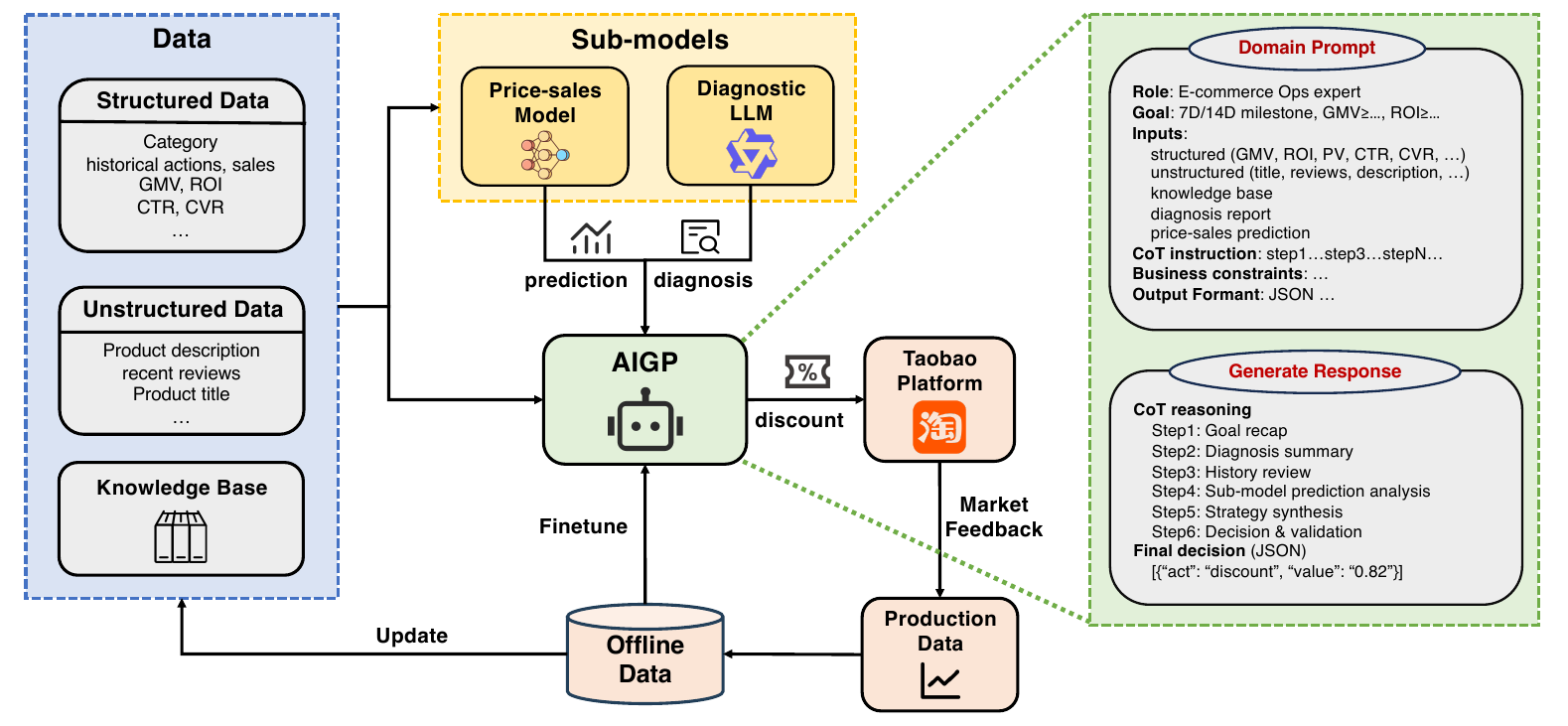}
\caption{Overall architecture of AIGP.}
\label{fig:overall}
\end{figure*}

\subsection{Preliminary}
Dynamic pricing in e-commerce is a sequential decision-making problem where pricing actions influence future market states and cumulative business outcomes. We define this at the SKU level, where the agent determines product prices based on market dynamics. To ensure ethical compliance, our framework is identity-blind, excluding individual user profiles to preclude price discrimination.

We formulate this problem as a Markov decision process (MDP) $\mathcal{M}=(\mathcal{S}, \mathcal{A}, P, r, \gamma)$, where $\mathcal{S}$ denotes the state space, $\mathcal{A}$ is the action space, $P\left(s_{t+1} \mid s_t, a_t\right)$ specifies transition dynamics, $r: \mathcal{S} \times \mathcal{A} \rightarrow \mathbb{R}$ is the reward function, and $\gamma \in(0,1)$ is the discount factor. At timestep $t$, the agent observes $x_t=\phi\left(s_t, c_t\right)$ encoding state $s_t$ and relevant task context $c_t$ (business goals and operational constraints), and selects a pricing action $a_t$ from the admissible set $\mathcal{A}_{\mathrm{safe}}\left(s_t\right)$ according to business and market rules. The objective is to maximize the expected cumulative discounted reward:
\begin{equation}
J(\pi) = \mathbb{E}_{\tau \sim \pi}\left[\sum_{t=0}^{T-1} \gamma^t r_t\right],
\end{equation}
where $\tau = (s_{0:T-1}, a_{0:T-1}, r_{0:T-1})$ is a trajectory generated by policy $\pi$.

In reinforcement learning, long-term value is modeled via the state-value function $V^\pi(s) = \mathbb{E}_{\pi}[\sum_{t=0}^{\infty} \gamma^t r_{t+1} \mid s_0 = s]$ and action-value function $Q^\pi(s, a) = \mathbb{E}_{\pi}[\sum_{t=0}^{\infty} \gamma^t r_{t+1} \mid s_0 = s, a_0 = a]$, satisfying the Bellman equations and are central to subsequent Long-Term Value Estimator (LTVE) modeling.

In this work, the pricing policy is parameterized by an LLM: $\pi_\theta(a\mid x) = \Psi(\operatorname{LLM}_\theta(x))$, where $\operatorname{LLM}_\theta(x)$ produces chain-of-thought reasoning and a structured decision, and $\Psi$ extracts the pricing action ensuring $a \in \mathcal{A}_{\mathrm{safe}}$ (definition in Appendix~\ref{appendix:a_safe_definition}). Unlike traditional RL policies outputting only actions, the LLM also generates interpretable reasoning traces explaining decisions.

\subsection{Domain-Adaptive Supervised Fine-Tuning}
\label{sec:sft}

When deploying LLMs as pricing policies, the model generates outputs containing both chain-of-thought (CoT) reasoning processes and final pricing decisions. This dual-output nature enables separate optimization: Supervised fine-tuning (SFT)~\cite{ouyang2022instructgpt} focuses on improving reasoning quality, instruction-following, and format compliance, establishing a foundation for subsequent decision optimization. Reasoning quality affects interpretability and trust, while decision accuracy directly impacts business outcomes. The overall workflow is illustrated in Fig.~\ref{fig:overall}.

Training data consists of high-quality instruction–response pairs. Each prompt integrates structured product attributes (sales history, exposure and click metrics, historical pricing actions, competitor statistics), business goals (GMV targets, ROI constraints, milestone achievements), and unstructured context (product descriptions, user reviews, domain knowledge). The prompt guides the model to analyze diagnostic insights, incorporate price-sales predictions, apply domain rules from the e-commerce operations knowledge base, and execute stepwise CoT reasoning ending with a compliant JSON-formatted action (see the prompt module and CoT-flow on the right side of Fig.~\ref{fig:overall}).

To balance inference efficiency and deployment costs for large-scale production use, we select a 30B parameter student model (Qwen3-30B-A3B~\cite{yang2025qwen3technicalreport}) for deployment. To enhance the student model's capabilities, we adopt teacher-student distillation where high-quality demonstrations are generated by a larger 235B parameter teacher model (Qwen3-235B-A22B~\cite{yang2025qwen3technicalreport}), which produces more sophisticated reasoning traces and more accurate decisions. 
We design an LLM-as-Judge module to automatically evaluate and filter teacher-generated responses across multiple quality dimensions. The judge model (Qwen3-235B-A22B-Instruct~\cite{yang2025qwen3technicalreport}) assesses each candidate response on four dimensions: (1) \textbf{Data Accuracy} verifies correct interpretation of numerical features and business constraints; (2) \textbf{Content Completeness} evaluates coverage of relevant factors and application of domain rules; (3) \textbf{Reasoning Internal Coherence} ensures logical consistency within the CoT process; (4) \textbf{Reasoning-Decision Consistency} checks alignment between reasoning conclusions and final pricing actions. Only responses with perfect scores (5/5) on all dimensions and actions satisfying $\mathcal{A}_{\mathrm{safe}}$ constraints (Appendix~\ref{appendix:a_safe_definition}) are retained. Judge reliability is validated via 96.4\% agreement with expert annotations on 2000 held-out responses. This workflow is summarized in part (a) of Fig.~\ref{fig:sft_dpo}.

Filtered high-quality responses are used to fine-tune the student model via parameter-efficient Low-Rank Adaptation (LoRA)~\cite{hu2022lora}. Model outputs must end with a valid JSON action satisfying platform and business criteria. Model optimization minimizes the cross-entropy loss:
\begin{equation}
\mathcal{L}_{\mathrm{SFT}}(\theta) = -\sum_{t=0}^{|y^{\star}|-1}\log \pi_\theta(y_t^{\star} \mid x, y_{<t}^{\star}).
\end{equation}
We apply mixed-precision computation, gradient clipping, and early stopping for efficiency and stability. As a result, SFT yields a policy that consistently follows instructions, generates transparent reasoning traces, and produces compliant pricing actions. However, SFT alone does not guarantee that the final pricing decisions align with long-term business value; further policy optimization through preference-based alignment is addressed in Section~\ref{sec:dpo}.

\subsection{Long-Term Value Modeling}
Accurately evaluating the long-term business impact of dynamic pricing policies is crucial but challenging due to complex delayed market effects and lack of verifiable correctness criteria. Unlike domains such as mathematical reasoning, dynamic pricing lacks deterministic quality metrics, making standard Reinforcement Learning from Human Feedback~\cite{ouyang2022instructgpt,safe-rlhf} (RLHF) or Reinforcement Learning from Verifiable Rewards~\cite{guo2025deepseek} (RLVR) difficult to apply. Base LLMs also lack domain-specific business knowledge and alignment with platform objectives.

To address this, we introduce a \textbf{Long-Term Value Estimator (LTVE)} $Q_\phi$, trained via offline reinforcement learning on over 5 million historical transitions from 6 months of production logs (60\% expert trajectories selected as top 30\% by cumulative GMV within comparable product groups, 40\% diverse non-expert cases for distribution coverage), to estimate expected cumulative reward for action $a$ in state $s$ and enable automatic preference pair construction for Direct Preference Optimization (Section~\ref{sec:dpo}).

\paragraph{\textbf{State and Action Spaces}}
Each state $s$ consists of structured features, including product attributes, sales statistics, inventory, historical ROI, engagement metrics (click-through rate, conversion rate), and temporal features such as day of week and promotion indicators.
The pricing action $a_t = d_t - d_{t-1}$ represents the daily discount rate change rather than the absolute discount level $d_t$. This design normalizes across products with heterogeneous baseline discount levels: different categories naturally operate at different price points, and modeling incremental changes allows the policy to learn generalizable adjustment strategies independent of absolute levels. Actions are constrained by business rules for feasible and safe adjustments (definition in Appendix~\ref{appendix:a_safe_definition}).

\paragraph{\textbf{Reward Design}}
The reward function balances long-term milestone achievement (tier thresholds based on 14-day cumulative GMV) and maintaining healthy ROI. To enable effective offline RL training, we adopt a \textbf{category-normalized relative reward}:
\begin{equation}
r_t = \lambda_1 (\mathrm{prog}_t - \mathrm{prog}_t^{\text{ref}}) + \lambda_2 (\log(1 + \mathrm{ROI}_t) - \log(1 + \mathrm{ROI}_t^{\text{ref}})),
\label{eq:reward_def}
\end{equation}
where $\mathrm{prog}_t$ measures daily contribution toward milestone achievement, and reference values $\mathrm{prog}_t^{\text{ref}}$, $\mathrm{ROI}_t^{\text{ref}}$ are category averages from products that successfully reached milestones within 14 days. This relative formulation isolates pricing strategy quality from intrinsic product characteristics: raw GMV and ROI are heavily influenced by category demand and brand recognition beyond the agent's control, and normalizing against category-level references ensures the reward signal reflects pricing decisions rather than product popularity. Log-transformation handles ROI's heavy-tailed distribution, and coefficients $\lambda_1, \lambda_2$ balance growth velocity and profitability.

\paragraph{\textbf{Offline RL Training}}
In real-world e-commerce, online exploration is highly constrained due to business risks and the lack of reliable simulators. Therefore, we train the Long-Term Value Estimator (LTVE) using offline RL on logged historical trajectories $\mathcal{D}= \{(s_t, a_t, r_t, \ldots)\}$. 
A critical challenge in e-commerce dynamic pricing is \textbf{delayed reward propagation}: pricing decisions on day $t$ affect not only immediate product performance, but also influence future exposure, traffic, and sales over multiple days through search and recommendation systems.

To model such delayed rewards in an offline and risk-controlled manner, we adopt a \textbf{critic-only architecture with multi-step temporal-difference (TD) targets}, learning action values directly from historical data without explicit policy training. We use a critic-only design because LTVE serves to rank candidate actions from a business-constrained set $\mathcal{A}_{\mathrm{safe}}(s)$ rather than generate unconstrained actions. Unlike policy-gradient methods that require broad action coverage, value-based ranking only needs accurate relative ordering within $\mathcal{A}_{\mathrm{safe}}(s)$, which is well-covered in historical logs.

We adopt a decoupled value learning framework~\cite{mnih2016asynchronous,wang2016dueling,iql} to enhance training stability and mitigate extrapolation errors. We separate the estimation of the state-value function $V_\psi(s)$ from the action-value functions $Q_{\phi_1}, Q_{\phi_2}$. The state-value network $V_\psi(s)$ is trained via expectile regression to approximate the upper quantiles of the action-value distribution:
\begin{equation}
\mathcal{L}_V(\psi)=
\mathbb{E}_{(s,a)\sim\mathcal{D}}
\left[
L_\tau\!\left(q_{\min}(s,a)-V_\psi(s)\right)
\right],
\label{eq:v_loss}
\end{equation}
where $L_\tau(u)=|\tau-\mathbb{I}(u<0)|u^2$ is the expectile loss with $\tau>0.5$, encouraging $V_\psi$ to generalize toward the best-performing actions without explicit action sampling. To mitigate overestimation bias, we use a clipped double-critic ensemble $q_{\min}=\min\!\left(Q_{\bar{\phi}_1},\,Q_{\bar{\phi}_2}\right)$.

The Q-critics are trained with multi-step Temporal Difference (TD) targets bootstrapped directly from $V_\psi$:
\begin{equation}
\mathcal{L}_Q(\phi_i)=
\mathbb{E}_{\mathcal{D}}
\left[
\left(Q_{\phi_i}(s_t,a_t)-y_t^{(n)}\right)^2
\right],\quad i\in\{1,2\},
\label{eq:q_loss}
\end{equation}
\begin{equation}
y_t^{(n)}=
\sum_{k=0}^{n-1}\gamma^k r_{t+k}+\gamma^n V_\psi(s_{t+n}).
\label{eq:nstep_target}
\end{equation}
Bootstrapping from $V_\psi$ alleviates the distribution shift problem by avoiding queries to out-of-distribution actions. Target critics are updated softly via $\bar{\phi}_i \leftarrow \eta \phi_i + (1-\eta)\bar{\phi}_i$ for $i\in\{1,2\}$, where $\eta$ is the soft update rate. For numerical stability, we clip $y_t^{(n)}$ to $[Q_{\min},Q_{\max}]$. The complete training procedure is summarized in Algorithm~\ref{alg:LTVE-core}. The trained LTVE serves as a long-term value scorer and facilitates the construction of preference pairs in Section~\ref{sec:dpo}.

\begin{algorithm}[t]
\caption{Long-Term Value Estimator (LTVE) Training}
\small
\label{alg:LTVE-core}
\begin{algorithmic}[1]
\Require Offline dataset $\mathcal{D}=\{(s_t,a_t,r_{t:t+n-1},s_{t+n})\}$
\Require Critics $Q_{\phi_1}, Q_{\phi_2}$ and target critics $Q_{\bar{\phi}_1}, Q_{\bar{\phi}_2}$
\Require Value network $V_\psi$
\Require Discount factor $\gamma$, expectile $\tau$, soft update rate $\eta$, clipping bounds $[Q_{\min}, Q_{\max}]$
\State Initialize $\phi_1,\phi_2,\psi$; set $\bar{\phi}_1\!\gets\!\phi_1$, $\bar{\phi}_2\!\gets\!\phi_2$
\For{each training iteration}
    \State Sample batch $\{(s_t,a_t,r_{t:t+n-1},s_{t+n})\}$ from $\mathcal{D}$
    \State $q_{\min} \gets \min(Q_{\bar{\phi}_1}(s_t,a_t),\, Q_{\bar{\phi}_2}(s_t,a_t))$
    \State $\delta \gets q_{\min} - V_\psi(s_t)$; \;\; $w \gets |\tau - \mathbb{I}(\delta < 0)|$
    \State Update $\psi$ by minimizing $\mathbb{E}\!\left[w\cdot \delta^2\right]$
    \State $y^{(n)} \gets \sum_{k=0}^{n-1}\gamma^k r_{t+k} + \gamma^n V_\psi(s_{t+n})$
    \State $y^{(n)} \gets \mathrm{clip}(y^{(n)}, Q_{\min}, Q_{\max})$ \Comment{target clipping}
    \State Update $\phi_1,\phi_2$ by minimizing $\mathbb{E}\!\left[(Q_{\phi_i}(s_t,a_t)-y^{(n)})^2\right]$
    \State $\bar{\phi}_1 \gets \eta \phi_1 + (1-\eta)\bar{\phi}_1$; \;\; $\bar{\phi}_2 \gets \eta \phi_2 + (1-\eta)\bar{\phi}_2$
\EndFor
\State \Return $Q_{\phi_1}, Q_{\phi_2}, V_\psi$
\end{algorithmic}
\end{algorithm}

\subsection{Preference Alignment with Long-Term Value Estimator}
\label{sec:dpo}
\begin{figure}[t]
\centering
\includegraphics[width=\linewidth]{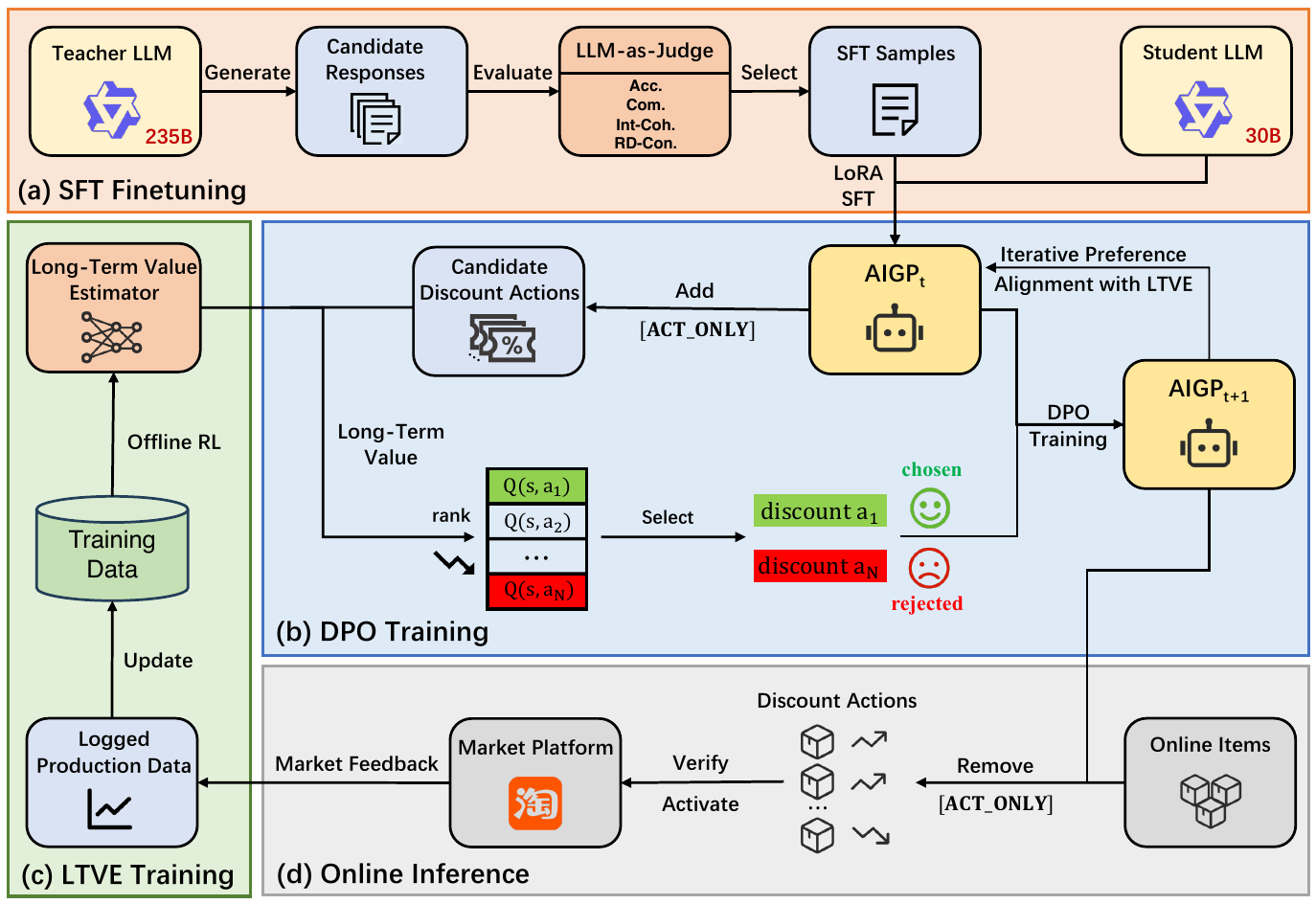}
\caption{The finetuning and inference pipeline of AIGP.}
\label{fig:sft_dpo}
\end{figure}

While supervised fine-tuning (SFT) improves instruction following, format compliance, and reasoning quality, it does not explicitly optimize pricing decisions for long-term business objectives. We therefore apply preference-based policy alignment with Direct Preference Optimization (DPO), guided by the Long-Term Value Estimator (LTVE). We use LTVE to curate preference data by ranking sampled candidate actions and selecting reliable chosen-rejected pairs. This reduces the impact of potential value extrapolation errors and provides a robust long-term alignment signal. The overall workflow is shown in Fig.~\ref{fig:sft_dpo}.

\paragraph{\textbf{Decision-only Mode for Preference Learning}}
To focus preference alignment on action selection, we append a \texttt{[ACT\_ONLY]} control token during candidate action generation and DPO training, so the model outputs only the structured discount action. This design addresses three challenges. First, constructing reliable preference pairs from full responses is difficult because reasoning quality is hard to compare and long-form text obscures the action's contribution to long-term value, the decision-only outputs allow LTVE to rank actions and generate preferences that directly reflect decision quality. Second, comparing full responses weakens the DPO signal with irrelevant tokens, whereas \texttt{[ACT\_ONLY]} concentrates optimization on the action tokens. Third, it avoids post-hoc action extraction, which can introduce distribution mismatch between preference pair construction and training. During deployment, the control token is removed so the model generates full reasoning traces for transparency. This workflow is illustrated in Fig.~\ref{fig:sft_dpo}(b).

\paragraph{\textbf{Candidate Action Sampling}}
For each prompt $x$ that encodes state $s$, we sample $M$ candidate actions
$\mathcal{A}(x)=\{a^{(1)},\ldots,a^{(M)}\}$
from the decision-only mode using stochastic decoding (temperature sampling with top-$k$ and top-$p$ filtering). Each candidate action is validated against business rules to ensure $a\in\mathcal{A}_{\mathrm{safe}}(s)$ (formal definition in Appendix~\ref{appendix:a_safe_definition}).

\paragraph{\textbf{Preference Pair Construction}}
We score each valid candidate action with LTVE to obtain $Q(s,a)$ as an estimate of the pricing action's long-term business value. Preference pairs are constructed by selecting two actions with sufficient separation in both value and discount magnitude:
\begin{equation}
Q(s,a^+) - Q(s,a^-) \ge \Delta_Q,
\qquad
|a^+ - a^-| \ge \Delta_a,
\end{equation}
where $a^+$ and $a^-$ denote the chosen and rejected pricing actions, and $\Delta_Q$ and $\Delta_a$ are thresholds controlling pair clarity. Since LTVE is only used to rank and filter candidates within the business-safe set, we further discard ambiguous pairs with small score gaps to reduce sensitivity to estimation noise.

\paragraph{\textbf{DPO Objective and Inference}}
Given a preference pair $(x,y^+,y^-)$, where $y^+$ and $y^-$ are the complete structured outputs (including JSON formatting) corresponding to chosen action $a^+$ and rejected action $a^-$, DPO minimizes:
\begin{equation}
\begin{split}
\mathcal{L}_{\mathrm{DPO}}(\pi_\theta;\pi_{\mathrm{ref}})
=
-\mathbb{E}_{(x,y^+,y^-)\sim\mathcal{D}}
\Big[
\log \sigma\Big(
\beta \Big(
\log \tfrac{\pi_\theta(y^+\mid x)}{\pi_{\mathrm{ref}}(y^+\mid x)}
\\
\qquad\qquad
-
\log \tfrac{\pi_\theta(y^-\mid x)}{\pi_{\mathrm{ref}}(y^-\mid x)}
\Big)
\Big)
\Big].
\end{split}
\label{eq:dpo}
\end{equation}
where $\pi_{\mathrm{ref}}$ is a frozen reference policy (the SFT model), and $\beta$ is a temperature parameter. Since \texttt{[ACT\_ONLY]} outputs only the pricing decision, the log-probabilities in Eq.~\eqref{eq:dpo} are computed on action tokens, making optimization focus on action selection. During training, \texttt{[ACT\_ONLY]} remains enabled to match the candidate generation distribution used for pair construction. For deployment, we remove the control token and prompt the aligned model to generate full reasoning traces together with the final structured action, preserving transparency and interpretability.

\paragraph{\textbf{Evaluation and Deployment}}
We evaluate the policy after preference alignment using offline metrics including LTVE score improvement, expert-action consistency, and rule compliance. Qualified models are deployed via large-scale online A/B testing, with real-time monitoring of cumulative GMV, integrated ROI, and long-term milestone completion against online production baselines.

\section{Experiment}
\subsection{Experimental Setup}

\subsubsection{Platform Deployment \& Data Description}
Experiments are conducted on Tao Factory, a vertical marketplace on Taobao with full pricing authority, focusing on optimizing daily discount rates for new products to accelerate sales growth and improve marketing ROI. The experimental dataset comprises millions of SKUs from over 180 days of production operations, including key information such as product attributes (category, brand, price, inventory), historical metrics (GMV, ROI, clic-through rate, conversion rate), temporal features, and unstructured context. The system makes offline decisions for about 200{,}000 products daily (actions effective next day). For offline evaluation, we employ a time-based split to prevent temporal data leakage. Our online A/B tests have been deployed and monitored for over 60 consecutive days, ensuring the sustained stability and effectiveness of the proposed framework.

\subsubsection{Implementation Details}
For LTVE, we train a critic-only double Q model using over 5 million transitions from 6 months of production logs, providing a robust foundation for long-term value estimation.
We set the discount factor $\gamma=0.95$, TD horizon $n=3$, and update soft update rate $\eta=0.01$. All LTVE networks are optimized with Adam (lr $=3\times 10^{-4}$).
For the LLM agent, SFT is conducted on 100{,}000 instruction-response pairs, and DPO uses 50{,}000 chosen-rejected preference pairs, both via LoRA adaptation on Qwen3-30B-A3B. Please refer to Appendix~\ref{appendix:extend:config} for complete implementation details.

\subsubsection{Baselines}
We compare AIGP against three categories of baselines: (1) \textbf{LLM Variants}, including base models and ablated versions to isolate the contribution of SFT, DPO, and the \texttt{[ACT\_ONLY]} control token; (2) \textbf{Online Deployed Policies}, representing production systems currently serving models; (3) \textbf{Academic RL Baselines}, trained on identical data and reward functions as LTVE for fair comparison. Brief descriptions are provided in Table~\ref{tab:baselines}, with comprehensive technical details in Appendix~\ref{appendix:extend:baseline}.

\begin{table}[t]
\centering
\caption{Brief descriptions of baseline methods.}
\label{tab:baselines}
\resizebox{\linewidth}{!}{
\begin{tabular}{ll}
\toprule
\textbf{Model} & \textbf{Brief Description} \\
\midrule
\multicolumn{2}{l}{\textbf{LLM Variants}} \\
Qwen3-30B-A3B~\cite{yang2025qwen3technicalreport} & Base Student LLM (30B params, no fine-tuning) \\
Qwen3-235B-A22B~\cite{yang2025qwen3technicalreport} & Base Teacher LLM (235B params, no fine-tuning) \\
AIGP (SFT-only) & Qwen3-30B-A3B with SFT only (w/o DPO) \\
AIGP (DPO-only) & Qwen3-30B-A3B with DPO only (base checkpoint as $\pi_{\text{ref}}$, w/o SFT) \\
AIGP (FullSeq) & AIGP (SFT+DPO) trained on full sequences (w/o \texttt{[ACT\_ONLY]}) \\
AIGP (SFT+DPO) & Qwen3-30B-A3B with SFT and DPO (full pipeline) \\
\midrule
\multicolumn{2}{l}{\textbf{Online Deployed Policies}} \\
Price-Sales Model & Deep Neural Network (DNN)-based Elasticity Model \\
RL-DT & Decision Transformer~\cite{chen2021decisiontransformer} for long-term goal optimization \\
\midrule
\multicolumn{2}{l}{\textbf{Academic RL Baselines}} \\
RL-DDPG~\cite{ddpg2019dynamic} & Deep Deterministic Policy Gradient \\
RL-SAC~\cite{sac2024dynamic} & Soft Actor-Critic with entropy regularization \\
\bottomrule
\end{tabular}
}
\end{table}

\subsection{Evaluation Metrics}
\subsubsection{Offline Evaluation Metrics}
\label{exp:metric:offline}
Offline evaluation covers decision quality (measured by Q-values and expert alignment) and reasoning quality (assessed by LLM-as-Judge).

For decision quality, we use Q-values from LTVE to measure the expected long-term business value of pricing decisions (reliability validated in Section~\ref{sec:eval_ltve}). Additionally, we use Expert Action Matching Accuracy (EAMA) to evaluate alignment with expert strategies:
\begin{equation}
\mathrm{EAMA}=\frac{1}{N} \sum_{i=1}^N \mathbb{I}\left(\left|\pi\left(s_i\right)-a_i^{\text{expert}}\right|<\epsilon\right),
\end{equation}
where $\pi(s_i)$ is the model's recommended discount, $a_i^{expert}$ is the expert action, and $\epsilon$ is a small threshold. 

For reasoning quality, we employ the LLM-as-Judge module described in Section~\ref{sec:sft} to evaluate model outputs on four dimensions: Data Accuracy (Acc.), Content Completeness (Com.), Reasoning Internal Coherence (Int-Coh.), and Reasoning-Decision Consistency (RD-Con.).

\subsubsection{Online Evaluation Metrics}
Business impact is evaluated over 7-day and 14-day horizons after product launch. Key metrics include cumulative Gross Merchandise Value (GMV), integrated Return on Investment (ROI), and Milestone Achievement Rate (MAR). MAR measures the proportion of SKUs reaching joint long-term GMV and ROI targets, ensuring immediate sales growth is balanced with financial sustainability. Significant gains in MAR demonstrate the model's capacity for stable long-term growth. 

\subsection{Evaluation of LTVE}
\label{sec:eval_ltve}
\subsubsection{Offline Evaluation Results}

We evaluate the reliability of the Long-Term Value Estimator (LTVE) on a held-out expert set of 30{,}000 samples. For operational fidelity, we use Expert Action Matching Accuracy (EAMA) as defined in Section~\ref{exp:metric:offline}. To measure the ability to distinguish optimal decisions, we compute Counterfactual Discrimination Accuracy (CDA):
\begin{equation}
\mathrm{CDA} = \frac{1}{N} \sum_{i=1}^{N} \mathbb{I}\left(Q(s_i, a_i^{\mathrm{expert}}) > Q(s_i, a_i^{\mathrm{cf}})\right),
\end{equation}
where $Q$ is the LTVE value function, $a_i^{expert}$ is the expert action, and $a_i^{cf}$ is a counterfactual variant.

\textbf{n-step TD ablation.} We ablate the multi-step TD horizon $n\in\{1,3,5\}$. As shown in Table~\ref{tab:ltve_eval}, $n=3$ achieves the best overall performance (MAE \textbf{0.027}, EAMA \textbf{90.7\%}, CDA \textbf{97.5\%}). We observe that $n=1$ provides insufficient credit assignment for delayed pricing effects in e-commerce, while $n=5$ amplifies off-policy estimation errors under limited action coverage in offline logs. The choice of $n=3$ strikes the best balance between capturing long-term impact and maintaining training stability. We thus use $n=3$ as the default.

\begin{table}[t]
\centering
\caption{Comprehensive evaluation of the LTVE with an ablation on the $n$-step TD horizon. Online A/B test results are reported for the deployed setting ($n=3$).}
\label{tab:ltve_eval}
\resizebox{\linewidth}{!}{
\begin{tabular}{llccc}
\toprule
\textbf{Evaluation Type} & \textbf{Metric} & \textbf{$n=1$} & \textbf{$n=3$} & \textbf{$n=5$} \\
\midrule
\multirow{3}{*}{Offline Evaluation} 
& MAE  & 0.045 & \textbf{0.027} & 0.056 \\
& EAMA ($\epsilon=0.05$) & 87.7\% & \textbf{90.7\%} & 86.4\% \\
& CDA  & 91.2\% & \textbf{97.5\%} & 89.7\% \\
\midrule
\multirow{3}{*}{Online A/B Test} 
& MAR & -- & \textbf{$+2.02\%$} & -- \\
& GMV & -- & \textbf{$+8.3\%$}  & -- \\
& ROI & -- & \textbf{$+7.2\%$}  & -- \\
\bottomrule
\end{tabular}
}
\end{table}

\begin{figure}[htbp]
  \centering
  \begin{minipage}[b]{0.7\linewidth}
    \centering
    \includegraphics[width=\linewidth]{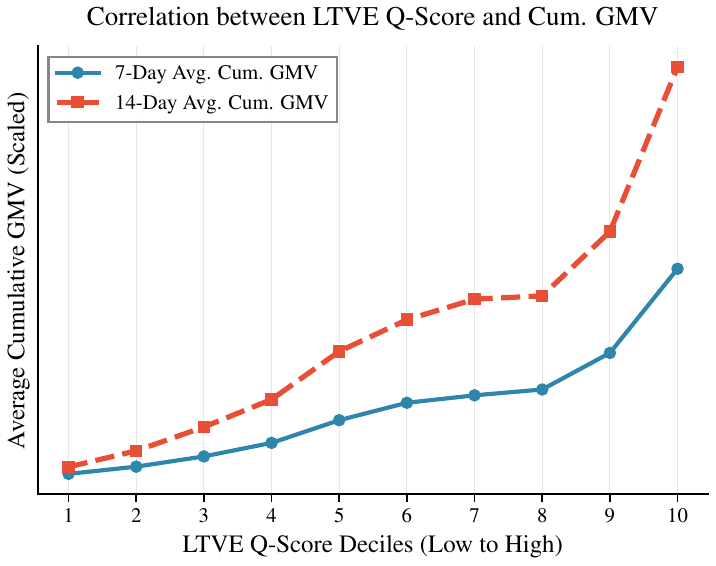}
  \end{minipage}
  \caption{Positive correlation between Q-score deciles and average cumulative GMV over 7-day and 14-day horizons.}
  \label{fig:q_func_effective}
\end{figure}

\textbf{Correlation with long-term outcomes.} 
To validate LTVE's predictive reliability, we partition 50,000 samples into 10 equal-sized $Q$-score deciles and track realized cumulative GMV over subsequent 7-day and 14-day periods, excluding the top and bottom 5\% of products within each decile to ensure robustness.
Fig.~\ref{fig:q_func_effective} shows strong monotonic correlation between $Q$-score deciles and realized GMV across both horizons.
Quantitatively, LTVE scores exhibit strong Spearman correlations with realized multi-objective outcomes: $\rho = 0.93/0.94$ with 7d/14d cumulative GMV, $\rho = 0.97/0.96$ with 7d/14d ROI, and $\rho = 0.97/0.98$ with 7d/14d milestone achievement rate (MAR).
These high correlations confirm that LTVE effectively captures long-term pricing consequences and enables reliable preference pair construction for policy alignment (Section~\ref{sec:dpo}).

\subsubsection{Online A/B Test}

We evaluate LTVE in online deployment by integrating it with a production Decision Transformer (RL-DT) model. For each product, RL-DT model generates multiple candidate discount actions. In the control group, DT's built-in critic selects the recommended discount. In the experimental group, LTVE replaces the critic and chooses the discount with the highest estimated long-term value. Products are randomly assigned to either group.

As shown in Table~\ref{tab:ltve_eval}, LTVE consistently improves business metrics over 7-day horizon: +2.02\% in Milestone Achievement Rate (MAR), +8.3\% in GMV, and +7.2\% in ROI, demonstrating effectiveness for real-world optimization.

\subsection{Evaluation of AIGP}
\subsubsection{Offline Evaluation Results}
We conduct comprehensive evaluations of AIGP across three dimensions: long-term value alignment, operational fidelity, and reasoning quality. To provide evaluation fully independent of LTVE, we introduce two return-weighted metrics that use only realized business outcomes as supervision. \textbf{EAMA-R} (Return-weighted EAMA) weights expert action matches by observed 3-day cumulative returns:
\begin{equation}
\mathrm{EAMA\text{-}R}=\frac{\sum_{i=1}^N \mathbb{I}\left(\left|\pi(s_i)-a_i^{\text{expert}}\right|<\epsilon\right) \cdot R_i^{3d}}{\sum_{i=1}^N R_i^{3d}},
\end{equation}
and \textbf{KIPS} (Kernel-weighted Industry Performance Score) replaces the hard threshold with Gaussian kernel proximity ($h{=}0.05$):
\begin{equation}
\mathrm{KIPS}=\frac{\sum_{i=1}^N \exp\!\left(-\frac{(\pi(s_i)-a_i^{\text{expert}})^2}{2h^2}\right) \cdot R_i^{3d}}{\sum_{i=1}^N R_i^{3d}},
\end{equation}
where $R_i^{3d} = \lambda_1 \cdot \mathrm{GMV}_i^{3d} + \lambda_2 \cdot \mathrm{ROI}_i^{3d}$ denotes the realized 3-day composite business return, computed entirely from observed platform metrics without any LTVE involvement. Higher values indicate stronger alignment with high-return expert decisions.

\paragraph{\textbf{Long-term Value Alignment}}
Table~\ref{tab:long_term_value_operational_fidelity} shows that full AIGP (SFT+DPO) achieves the highest Q-score of \textbf{2.836}, significantly exceeding production baselines (Price-Sales: 2.564; RL-DT: 2.694) and LLM variants. Academic RL baselines (RL-DDPG, RL-SAC) yield lower scores, demonstrating AIGP's superiority in aligning decisions with long-term business value.

\begin{table}[htbp]
\centering
\caption{Comparison of long-term Q-scores and Expert Trajectory Overlap metrics across model variants and baselines.}
\label{tab:long_term_value_operational_fidelity}
\resizebox{\linewidth}{!}{
\begin{tabular}{c|c|cc|cc}
\toprule
\textbf{Model} & \textbf{Avg. Q-Score} & \textbf{MAE} & \textbf{EAMA} & \textbf{EAMA-R} & \textbf{KIPS}\\
\midrule
Price-Sales         & 2.564 & 0.084   & 71.08\%  & 11.62 & 11.68  \\
RL-DT                       & 2.694 & 0.078   & 73.89\% & 11.65 & 11.71  \\
RL-DDPG                     & 2.364         & 0.110        & 56.91\%  & 11.41 & 11.47  \\
RL-SAC                      & 2.494         & 0.097        & 63.95\%  & 11.52 & 11.57  \\
Qwen3-30B-A3B   & 2.585 & 0.085   & 69.22\% & 11.74 & 11.80  \\
Qwen3-235B-A22B  & 2.556         & 0.089        & 67.07\%  & 11.69 & 11.76  \\
AIGP (SFT-only)    & 2.554 & 0.086   & 68.98\% & 11.71 & 11.77  \\
AIGP (DPO-only)    & 2.794         & 0.067        & 79.26\%  & 11.81 & 11.84  \\
AIGP (FullSeq) & 2.591 & 0.082 & 71.32\% & 11.73 & 11.78  \\
AIGP (SFT+DPO)         & \textbf{2.836} & \textbf{0.062} & \textbf{82.51\%} & \textbf{11.83} & \textbf{11.86} \\
\bottomrule
\end{tabular}
}
\end{table}

\paragraph{\textbf{Operational Fidelity}}
We assess operational fidelity by measuring alignment of actions with expert trajectories using Mean Absolute Error (MAE), Expert Action Matching Accuracy (EAMA with $\epsilon{=}0.05$) and the LTVE-independent metrics EAMA-R and KIPS. Table~\ref{tab:long_term_value_operational_fidelity} shows full AIGP pipeline achieves the lowest MAE (\textbf{0.062}), highest EAMA (\textbf{82.51\%}), EAMA-R (\textbf{11.83}), and KIPS (\textbf{11.86}). The consistent rankings across LTVE-dependent (Q-Score) and LTVE-independent (EAMA-R, KIPS) metrics confirm that performance gains are genuine rather than artifacts of the value estimator.

\begin{figure}[t]
  \centering
  \begin{minipage}[b]{0.9\linewidth}
    \centering
    \includegraphics[width=\linewidth]{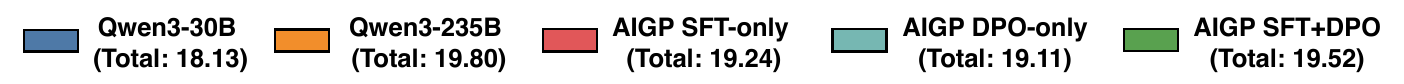}
  \end{minipage}
  \begin{minipage}[b]{0.9\linewidth}
    \centering
    \includegraphics[width=\linewidth]{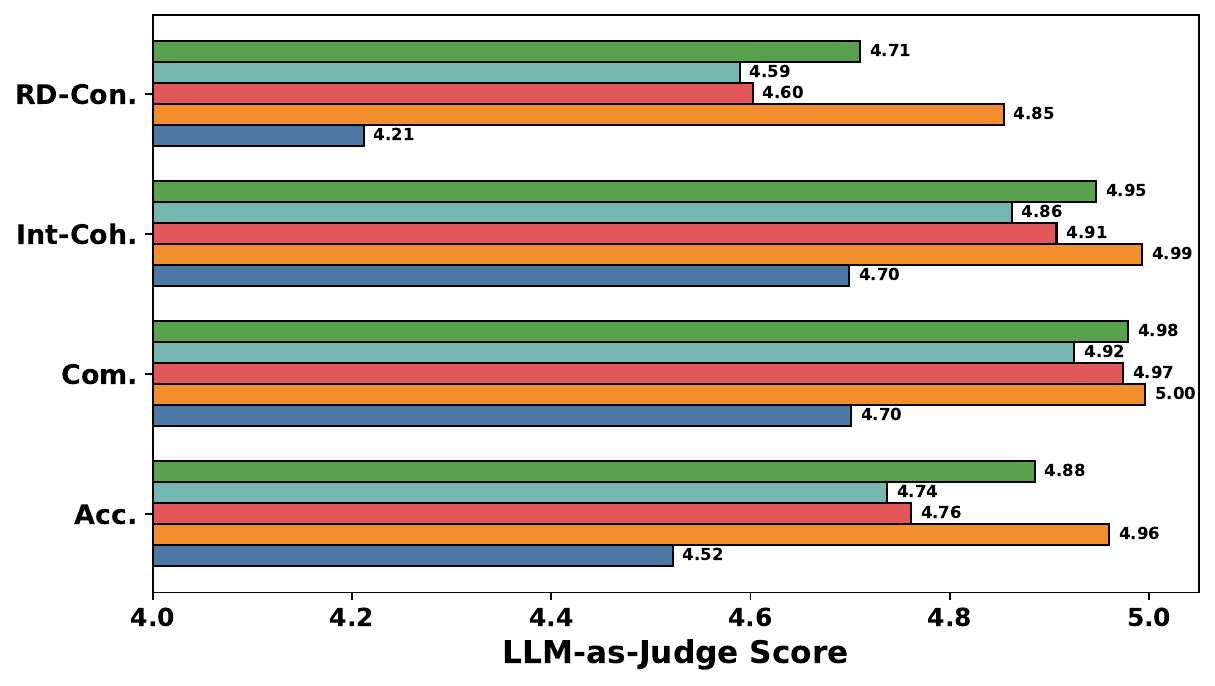}
  \end{minipage}
  \caption{LLM-as-judge scores of reasoning quality across model variants.}
  \label{fig:llm_as_judge:result}
\end{figure}

\paragraph{\textbf{Reasoning quality}}
Text generation quality is assessed via LLM-as-Judge rubric on data accuracy, content completeness, reasoning internal coherence (Int-Coh.), and reasoning-decision consistency (RD-Con.). As shown in Fig.~\ref{fig:llm_as_judge:result}, while Qwen3-235B scores highest (\textbf{19.80}), reflecting the well-established correlation between model size and generation quality as described by scaling laws~\cite{kaplan2020scaling,hoffmann2022training}, both SFT and DPO effectively enhance the reasoning performance of the smaller Qwen3-30B model. SFT alone improves the total score from 18.13 to 19.24, and combining SFT with DPO achieves \textbf{19.52}, approaching the 235B teacher model across all dimensions. This indicates that SFT and DPO facilitate domain-specific knowledge condensation, distilling specialized pricing heuristics into compact architectures. Consequently, targeted alignment significantly closes the size-induced performance gap, enabling deployable models to match the reasoning quality of much larger LLMs.

\begin{table*}[t]
\caption{Online A/B Test Results at 7-Day and 14-Day Horizons (relative improvements over Price-Sales Model).}
\label{tab:online:ab}
\centering
\small
\begin{tabular}{c|ccc|ccc}
\toprule
\textbf{Model} & \textbf{7-Day-MAR}    & \textbf{7-Day-GMV}     & \textbf{7-Day-ROI}     & \textbf{14-Day-MAR}     & \textbf{14-Day-GMV}      & \textbf{14-Day-ROI}     \\
\midrule
RL-DT           & $+1.33\%$ & $+4.02\%$ & $+5.71\%$ & $+0.88\%$  & $+9.04\%$  & $+3.46\%$  \\
Qwen3-30B-A3B   & $+0.88\%$ & $+2.22\%$ & $+2.29\%$ & $-2.98\%$  & $+4.89\%$  & $+4.33\%$  \\
Qwen3-235B-A22B & $+5.69\%$ & $+8.67\%$ & \textbf{+8.52\%} & $+7.37\%$  & $+10.50\%$ & $+6.62\%$  \\
AIGP (SFT-only) & $+5.59\%$ & $+4.82\%$ & $+8.00\%$ & $+3.34\%$  & $+7.85\%$ &  $+5.29\%$  \\
AIGP (SFT+DPO)  & \textbf{+6.92\%} & \textbf{+11.04\%} & $+8.23\%$ & \textbf{+8.20\%} & \textbf{+13.21\%} & \textbf{+7.59\%} \\
\bottomrule
\end{tabular}
\end{table*}

\subsubsection{Online A/B Test}
We conducted large-scale A/B tests on Tao Factory by deploying AIGP alongside baselines. To ensure rigorous evaluation, approximately 200,000 SKUs were randomly assigned to mutually exclusive groups via a Hash-based stable partitioning (Appendix~\ref{appendix:hash_partition}), maintaining policy consistency throughout each SKU's lifecycle. All groups were verified to be well-balanced across exposure, category, and price levels. The experiment ran continuously for over 60 days with daily monitoring to verify stability.

Performance is measured over 7-day and 14-day horizons on Milestone Achievement Rate (MAR), cumulative GMV, and integrated ROI. Table~\ref{tab:online:ab} reports relative improvements over the Price-Sales baseline. AIGP (SFT+DPO) achieves strong and sustained performance across all metrics. Over 7 days, AIGP achieves \textbf{+6.92\%} MAR, \textbf{+11.04\%} GMV, and \textbf{+8.23\%} ROI. Over 14 days, gains remain stable at \textbf{+8.20\%} MAR, \textbf{+13.21\%} GMV, and \textbf{+7.59\%} ROI. These results demonstrate that AIGP drives substantial business value and operational efficiency at scale.

\subsubsection{Pricing Adjustment Stability}
\label{exp:stability}

\begin{figure}[t]
    \centering
    \includegraphics[width=\linewidth]{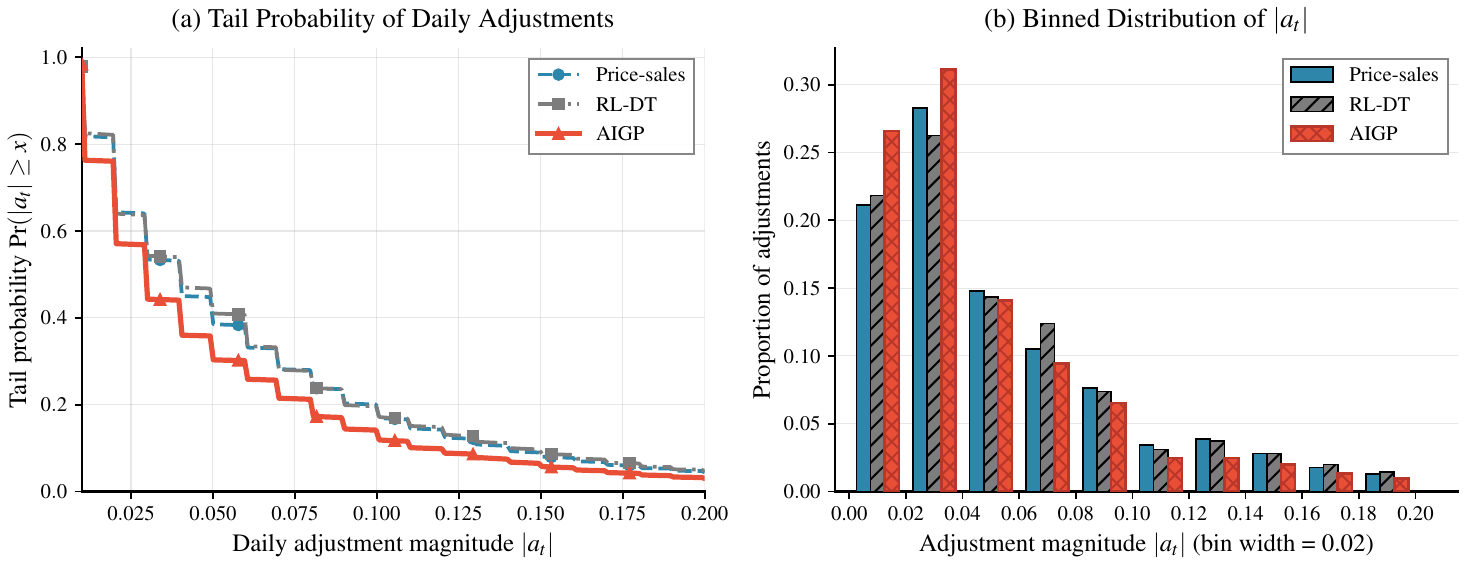}
    \caption{\textbf{Pricing adjustment stability over the first 14 days after launch.}
(a) Tail probability $\Pr(|a_t|\ge x)$.
(b) Binned distribution of $|a_t|$ (bin width = 0.02).}
    \label{fig:discount_adjustment_profile}
\end{figure}

Beyond business outcomes, we analyze daily discount adjustments $a_t=d_t-d_{t-1}$ to verify AIGP's pricing stability, which is critical for new products where traditional models often exhibit high pricing volatility. 
Fig.~\ref{fig:discount_adjustment_profile} illustrates the distribution of daily adjustments over the first 14 days after launch.
Fig.~\ref{fig:discount_adjustment_profile} (a) shows AIGP yields lower tail probabilities at larger thresholds, indicating fewer extreme discount jumps compared to production baselines. Furthermore, Fig.~\ref{fig:discount_adjustment_profile} (b) reveals that AIGP assigns a higher proportion of adjustments to small-magnitude bins (e.g., $|a_t|<0.04$), demonstrating a more fine-grained and controlled update pattern.
This pricing stability, achieved by integrating unstructured context and analogical reasoning, avoids abrupt price swings while providing merchants with predictable dynamics for inventory planning. Moreover, AIGP's explicit natural-language rationales enable post-hoc auditing, supporting safer large-scale deployment.


\subsubsection{Ablation Studies}
\label{exp:ablation:dpo}
We ablate the core components of AIGP across five LLM variants: Qwen3-30B base, SFT-only, DPO-only, AIGP (FullSeq), and full AIGP.
As shown in Table~\ref{tab:long_term_value_operational_fidelity}, SFT primarily enhances reasoning quality and output consistency, whereas DPO is critical for boosting long-term value alignment and expert matching accuracy. Specifically, DPO-only achieves higher Q-score, EAMA, and KIPS (2.794, 79.26\%, 11.84) compared to SFT-only (2.554, 68.98\%, 11.77), with consistent gains across all LTVE-independent metrics.

To verify the necessity of the \texttt{[ACT\_ONLY]} control token, we evaluate the AIGP (FullSeq) variant, which is trained using DPO on the complete output sequences (including both reasoning traces and pricing actions) rather than the action-only mode. The results indicate that without the control token, the Q-Score (2.591) and EAMA (71.32\%) only marginally outperform the SFT-only baseline and fall significantly short of the full AIGP (2.836). This performance gap confirms that when the pricing action is embedded within long reasoning traces, the DPO signal becomes heavily diluted, making it difficult for the model to isolate and optimize the specific decision tokens.

Combining SFT and DPO with the \texttt{[ACT\_ONLY]} token delivers the best overall performance across reasoning quality (Judge score 19.52), long-term value (Q-value 2.836), and operational fidelity (EAMA 82.51\%). Online testing (Table~\ref{tab:online:ab}) further confirms that full AIGP achieves superior business outcomes compared to all ablated variants. These results highlight that while SFT enables interpretable reasoning, the \texttt{[ACT\_ONLY]} guided DPO is essential for optimizing long-term outcomes in complex e-commerce scenarios.

\paragraph{\textbf{Impact of Unstructured Inputs}}
To quantify the contribution of unstructured context, we ablate two key sources: the product diagnosis report (Diag) and the domain knowledge base (KB). As shown in Table~\ref{tab:ablation_unstructured}, removing either source causes consistent degradation across all metrics and model variants. For the full AIGP (SFT+DPO), removing KB leads to a Q-score drop of 0.236 and KIPS drop of 0.13, while removing Diag causes smaller but consistent declines (Q-score -0.176, KIPS -0.11). KB removal has a uniformly larger impact, indicating that domain knowledge provides critical pricing priors especially for cold-start products. These results confirm that although LTVE operates solely on structured features, the LLM's reasoning over unstructured inputs substantially improves decision quality.

\begin{table}[t]
\centering
\small
\caption{Ablation on unstructured inputs: Diagnosis report (Diag) and Knowledge base (KB).}
\label{tab:ablation_unstructured}
\begin{tabular}{c|c|cc}
\toprule
\textbf{Model} & \textbf{Avg. Q-Score} & \textbf{EAMA-R} & \textbf{KIPS}\\
\midrule
AIGP (SFT-only)      & 2.554 & 11.71 & 11.77 \\
\quad w/o Diag       & 2.470 {\scriptsize(-0.084)} & 11.66 {\scriptsize(-0.05)} & 11.73 {\scriptsize(-0.04)} \\
\quad w/o KB         & 2.330 {\scriptsize(-0.224)} & 11.56 {\scriptsize(-0.15)} & 11.63 {\scriptsize(-0.14)} \\
\midrule
AIGP (DPO-only)      & 2.794 & 11.81 & 11.84 \\
\quad w/o Diag       & 2.680 {\scriptsize(-0.114)} & 11.68 {\scriptsize(-0.13)} & 11.75 {\scriptsize(-0.09)} \\
\quad w/o KB         & 2.480 {\scriptsize(-0.314)} & 11.57 {\scriptsize(-0.24)} & 11.69 {\scriptsize(-0.15)} \\
\midrule
AIGP (SFT+DPO)      & \textbf{2.836} & \textbf{11.83} & \textbf{11.86} \\
\quad w/o Diag       & 2.660 {\scriptsize(-0.176)} & 11.67 {\scriptsize(-0.16)} & 11.75 {\scriptsize(-0.11)} \\
\quad w/o KB         & 2.600 {\scriptsize(-0.236)} & 11.59 {\scriptsize(-0.24)} & 11.73 {\scriptsize(-0.13)} \\
\bottomrule
\end{tabular}
\end{table}

\paragraph{\textbf{Comparison with Strong Open-Source LLMs}}
To contextualize AIGP's gains, we compare against recent open-source LLMs deployed on-premise under identical prompts and evaluation protocols. Due to data sensitivity (proprietary pricing/sales data), external closed-source APIs are excluded. As shown in Table~\ref{tab:sota_llm}, stronger base models (Qwen3.5-35B) perform competitively but still fall short of AIGP's aligned pipeline. General-purpose reasoning models (DeepSeek-R1~\cite{guo2025deepseek}) do not transfer well to domain-specific pricing, underscoring the necessity of task-specific alignment over scaling alone.

\begin{table}[t]
\centering
\small
\caption{Comparison with other strong open-source LLMs.}
\label{tab:sota_llm}
\begin{tabular}{c|c|cc}
\toprule
\textbf{Model} & \textbf{Avg. Q-Score} & \textbf{EAMA-R} & \textbf{KIPS}\\
\midrule
Qwen3-30B-A3B       & 2.585 & 11.74 & 11.80 \\
Qwen3.5-35B-A3B     & 2.730 & 11.75 & 11.81 \\
DeepSeek-R1-32B     & 2.550 & 11.72 & 11.78 \\
DeepSeek-R1-14B     & 2.450 & 11.71 & 11.76 \\
\midrule
AIGP (SFT+DPO)      & \textbf{2.836} & \textbf{11.83} & \textbf{11.86} \\
\bottomrule
\end{tabular}
\end{table}

\subsubsection{Robustness Analysis}
\label{exp:robustness}
To identify potential fragility of LTVE and evaluate how AIGP (SFT+DPO) compensates, we measure Counterfactual Discrimination Accuracy (CDA) across challenging scenarios, comparing against Conservative Q-Learning (CQL)~\cite{kumar2020cql}. As shown in Table~\ref{tab:ltve_scenario}, two primary fragility scenarios emerge from out-of-distribution (OOD) challenges: (1) \textit{Cold-start} products with limited history degrade both LTVE and CQL, but AIGP (SFT+DPO) compensates via LLM reasoning over unstructured context (94.6\%); (2) \textit{Boundary actions} (large adjustments rarely seen in logs) where LTVE's dual-critic clipping mitigates OOD suppression, and AIGP (SFT+DPO) further improves through generalization (94.1\%).

\begin{table}[t]
\centering
\small
\caption{Counterfactual Discrimination Accuracy across scenarios. AIGP (SFT+DPO) combines LTVE with LLM reasoning.}
\label{tab:ltve_scenario}
\begin{tabular}{c|ccc}
\toprule
\textbf{Scenario} & \textbf{LTVE} & \textbf{CQL} & \textbf{AIGP (SFT+DPO)}\\
\midrule
Overall              & \textbf{97.5\%} & 93.1\% & 96.9\% \\
Cold-start (on-shelf $\leq$3 days) & 89.3\% & 83.5\% & \textbf{94.6\%} \\
Boundary action ($|a_t|\geq$0.15) & 91.2\% & 80.7\% & \textbf{94.1\%} \\\bottomrule
\end{tabular}
\end{table}

\subsubsection{Case Study}
\label{exp:case_study}
\begin{figure}[t]
    \centering
    \includegraphics[width=\linewidth]{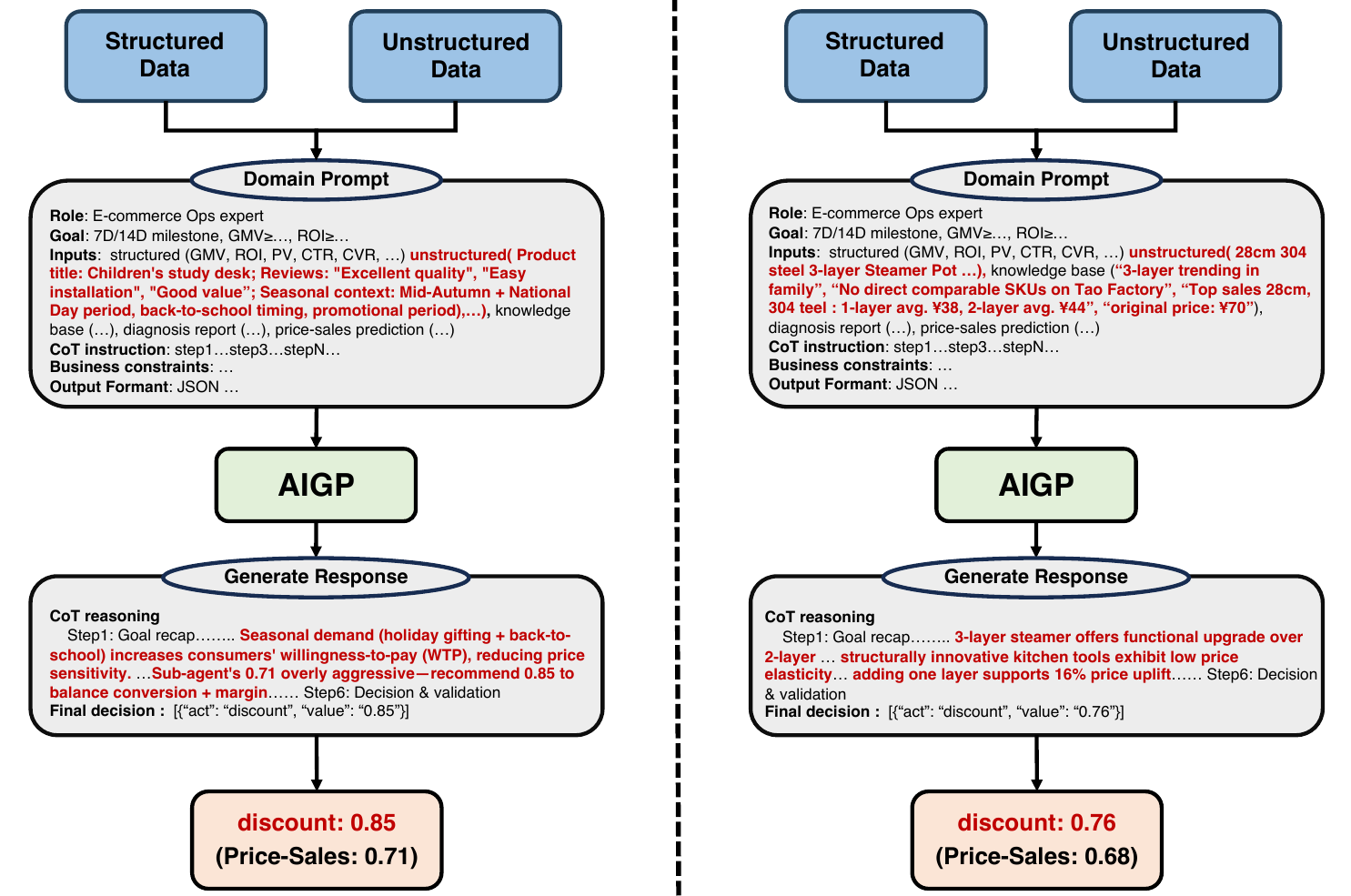}
    \caption{AIGP compared to traditional models. (Left) Case 1: Utilizing unstructured information; (Right) Case 2: Cold-start adaptation.}
    \label{fig:case_study}
\end{figure}

We present two scenarios demonstrating AIGP's advantages: (1) leveraging unstructured information, and (2) cold-start generalization. As shown in Fig.~\ref{fig:case_study}, \textbf{Case 1} shows AIGP pricing a children's desk by integrating reviews (Excellent quality) and seasonal context (back-to-school). CoT reasoning identifies that seasonal demand increases willingness-to-pay while positive reviews reduce price sensitivity. Unlike traditional price-sales models that rely solely on historical transaction density and often over-discount during peak seasons, AIGP recognizes the intrinsic value signaled by textual feedback. This allows it to yield a discount rate of 0.85 compared to the baseline's 0.71, successfully preserving 14\% margin without sacrificing sales velocity. \textbf{Case 2} shows AIGP handling a cold-start 3-layer steamer with minimal historical data. AIGP leverages its internal knowledge base, which contains pricing benchmarks for related products such as 1-layer (¥38) and 2-layer (¥44) steamers. It applies analogical reasoning to recommend a more balanced discount rate of 0.76 versus baseline's 0.68. While black-box RL models often exhibit high volatility or conservative bias in data-sparse scenarios, AIGP's reasoning module bridges the gap via logical extrapolation from category-level heuristics.

Both cases highlight three core strengths: 
(1) \textbf{Unstructured data utilization}, where AIGP converts qualitative signals into quantitative pricing. 
(2) \textbf{Cold-start robustness}, which overcomes data sparsity through analogical knowledge transfer. 
(3) \textbf{Interpretability}, provided via explicit CoT traces that justify decisions to ensure merchant trust and auditable deployment.

\section{Conclusion \& Future Work}
We propose \textbf{AIGP}, a novel framework integrating LLM-based reasoning with offline reinforcement learning for interpretable and business-aligned dynamic pricing. By combining supervised fine-tuning with Direct Preference Optimization guided by a Long-Term Value Estimator (LTVE), AIGP facilitates a synergy between generative reasoning and long-term business value optimization. This approach enables domain-specific knowledge condensation, allowing compact, deployable models to achieve decision quality and reasoning depth competitive with much larger LLMs while ensuring policy transparency. Extensive evaluations on Tao Factory demonstrate that AIGP significantly outperforms established baselines in cumulative GMV and milestone achievement, validating its ability to drive sustainable growth at scale.

Future work will pursue two directions: (1) exploring the cross-domain extensibility of the AIGP framework to broader e-commerce decision tasks; and (2) refining alignment strategies to develop a unified, efficient framework that simultaneously enhances symbolic reasoning and strategic decision-making capabilities.

\bibliographystyle{ACM-Reference-Format}
\balance
\bibliography{cite}

\appendix

\section{Extended Experimental Configuration}
\label{appendix:extend:config}

\subsection{Detailed Hyper-parameters}

Table~\ref{tab:appendix:hyperparams} lists the detailed configurations for the LTVE training, SFT, and DPO alignment stages.

\begin{table}[htbp] 
  \centering
  \caption{Comprehensive Hyper-parameters for AIGP.}
  \label{tab:appendix:hyperparams}
  \footnotesize
  \begin{tabular}{lp{4.8cm}}
    \toprule
    \textbf{Category} & \textbf{Parameter / Value} \\
    \midrule
    \multicolumn{2}{l}{\textit{Long-Term Value Estimator (LTVE)}} \\
    \midrule
    Discount Factor $\gamma$ & 0.95 \\
    $n$-step Horizon & 3 \\
    Expectile $\tau$ & 0.8 \\
    Learning Rate & $3 \times 10^{-4}$ (Adam) \\
    Soft Update Rate $\eta$ & 0.01 \\
    Batch Size & 512 \\
    Reward Milestone Weight $\lambda_1$ & 0.8 \\
    Reward ROI Weight $\lambda_2$ & 0.2 \\
    \midrule
    \multicolumn{2}{l}{\textit{LLM Supervised Fine-Tuning (SFT)}} \\
    \midrule
    Learning Rate & $5 \times 10^{-5}$ (Cosine Scheduler) \\
    LoRA Target & o\_proj, q\_proj, k\_proj, v\_proj \\
    Cutoff Length & 8192 \\
    Epochs & 2 \\
    \midrule
    \multicolumn{2}{l}{\textit{LLM Preference Alignment (DPO)}} \\
    \midrule
    Learning Rate & $1 \times 10^{-5}$ \\
    DPO $\beta$ & 0.1 \\
    LoRA Target & o\_proj, q\_proj, k\_proj, v\_proj \\
    Cutoff Length & 4096 \\
    Epochs & 3 \\
    \bottomrule
  \end{tabular}
\end{table}

\subsection{Baseline Implementation and Descriptions}
\label{appendix:extend:baseline}

\textbf{Price-Sales Model.} This is the current production baseline deployed online. It utilizes a deep neural network to model price-sales elasticity, specifically predicting the uplift of sales relative to price adjustments. However, it lacks the ability to incorporate unstructured context (like user reviews) and fails to optimize for long-term objectives such as milestone achievement. Due to commercial confidentiality, further technical specifics of the neural architecture are omitted, but it represents a high-performing point-estimate approach widely used in large-scale e-commerce.

\textbf{RL-DT (Decision Transformer).} RL-DT model the daily dynamic pricing task as a sequence-to-sequence decision-making problem via Decision Transformer~\cite{chen2021decisiontransformer} architecture. Unlike traditional RL, RL-DT leverages the transformer's ability to process long-range historical sequences, predicting the next pricing action conditioned on the Return-to-go (the expected cumulative reward). This makes it a strong baseline for long-term goal optimization. The model uses the same historical transitions and reward signals as our framework to ensure comparability.

\textbf{RL-DDPG.} We implement a continuous control baseline based on the Deep Deterministic Policy Gradient~\cite{ddpg2019dynamic} algorithm. It follows an Actor-Critic architecture where the actor network outputs deterministic discount rates and the critic estimates the Q-value. To ensure a fair comparison, the dataset and reward construction are strictly identical to those used for training our LTVE.

\textbf{RL-SAC.} We also adapt the Soft Actor-Critic~\cite{sac2024dynamic} algorithm, which incorporates an entropy-regularized Actor-Critic framework. This baseline is designed to improve robustness and exploration by maximizing both the expected reward and the policy entropy. To ensure a fair comparison, the dataset and reward construction are strictly identical to those used for training our LTVE.

\textbf{Base LLMs.} We include base Qwen3-30B-A3B and Qwen3-235B-A22B~\cite{yang2025qwen3technicalreport} models that receive the same structured and unstructured inputs but make pricing decisions based solely on their pre-trained general knowledge. This baseline quantifies the specific gains achieved through our long-term value alignment and supervised domain adaptation.

\section{Deployment Details of Online A/B Test}
\subsection{Hash-based Partitioning}
\label{appendix:hash_partition}
To ensure the statistical validity of our online A/B tests and maintain policy consistency for each product, we implement a \textit{Hash-based Stable Partitioning} mechanism. The core objective is to map each Stock Keeping Unit (SKU) into mutually exclusive experimental or control groups based on its unique identifier, ensuring that approximately 200,000 SKUs are assigned consistently throughout the testing period. The partitioning process follows these technical steps:

\textbf{Identifier Selection.} We use the unique \texttt{SKU\_ID} as the primary key for partitioning to ensure that the same product is always processed by the same pricing policy throughout its lifecycle.

\textbf{Salted Hashing.} To prevent potential bucket coupling (where products are always grouped together across different independent experiments), we append a specific \texttt{Experiment\_Salt} to the \texttt{SKU\_ID} before hashing. This salted approach also mitigates selection bias that might arise from semi-semantic \texttt{SKU\_ID} structures by ensuring high entropy in the hash space.

\textbf{Modulo Mapping.} The salted ID is processed using a cryptographic hash function, and the resulting integer is mapped to a set of discrete buckets via a modulo operation. This mapping ensures a deterministic and uniform distribution of SKUs across experimental cohorts.

\textbf{Group Assignment.} Based on pre-defined traffic allocation ratios, SKUs falling into specific numeric buckets are assigned to their respective treatment groups, such as the AIGP (SFT+DPO) cohort or the production baseline.

This method guarantees \textbf{policy stability}. Since the platform makes daily decisions for a vast number of products, any fluctuation in group assignment would lead to erratic pricing signals and destabilize merchant inventory planning. By using Hash-based Partitioning, we ensure that experimental groups remain well-balanced across exposure, category, and price levels, thereby isolating the true business impact of the AIGP framework.

\subsection{Safety and Compliance Mechanisms}
\label{appendix:a_safe_definition}
To ensure business stability and regulatory compliance during online deployment, we formally define the admissible action space $\mathcal{A}_{\mathrm{safe}}(s_t)$. In our framework, a pricing action $a_t$ represents the daily change in the discount rate, i.e., $a_t = d_t - d_{t-1}$, where $d_t$ is the absolute discount rate applied at time $t$. 

The safety space $\mathcal{A}_{\mathrm{safe}}$ is governed by the following operational constraints:

\textbf{Daily Adjustment Constraints.} To prevent abrupt price swings that may destabilize market expectations or merchant inventory planning, the daily adjustment magnitude is restricted to a fixed range:
\begin{equation}
    a_t \in [-0.2, 0.2]
\end{equation}
This ensures that the discount rate cannot fluctuate by more than 20\% within a single 24-hour cycle.

\textbf{Absolute Discount Boundaries.} To maintain brand value and protect minimum profit margins, the resultant absolute discount rate $d_t$ must remain within a predefined sustainable interval:
\begin{equation}
    d_t \in [0.5, 1.0]
\end{equation}
where $d_t = 1.0$ represents the original listing price. Any discount exceeding 50\% ($d_t < 0.5$) is prohibited to prevent erosion of Return on Investment (ROI).

Formally, given the previous state's discount $d_{t-1}$, the set of admissible actions at time $t$ is defined as:
\begin{equation}
    \mathcal{A}_{\mathrm{safe}}(s_t) = \{ a_t \mid -0.2 \leq a_t \leq 0.2 \text{ and } 0.5 \leq d_{t-1} + a_t \leq 1.0 \}
\end{equation}

As illustrated in our inference pipeline in Fig~\ref{fig:sft_dpo}(d), any action $a_t$ generated by the pricing policy that violates these boundaries is projected back to the nearest feasible point in $\mathcal{A}_{\mathrm{safe}}$ before platform execution. This mechanism guarantees that the agent's exploration remains within the safe operational manifold of the e-commerce platform.

\subsection{Infrastructure and Deployment Efficiency}
\label{appendix:infrastructure}

\textbf{Hardware Constraints and Model Selection.} 
To ensure the practical feasibility of the AIGP framework in a large-scale production environment, we conducted all model alignment and experimental evaluations using a cluster of 16 GPUs. While the 235B teacher model exhibits superior general-purpose reasoning, its parameter scale poses prohibitive computational demands for fine-tuning within our available infrastructure. Specifically, the 16-GPU cluster is insufficient to support even parameter-efficient fine-tuning (e.g., LoRA) for a 235B-scale model due to severe memory bottlenecks and excessive training latency. In contrast, the 30B student model (Qwen3-30B-A3B) allows for high-quality SFT and DPO alignment under these hardware constraints, effectively internalizing domain-specific pricing knowledge.

\textbf{Operational Pipeline and Inference Capacity.} 
The AIGP system serves a vertical marketplace with approximately 200,000 active SKUs per day. Instead of real-time price adjustments, our framework follows a daily offline execution pipeline. Decisions are computed in batch during nightly processing windows, with the resulting pricing strategies synchronized to the production environment for execution on the subsequent day (T+1). This operational design prioritizes total throughput for massive SKU volumes over immediate response latency. Utilizing the 30B model significantly optimizes this pipeline:
\begin{itemize}
    \item \textbf{Total Inference Duration.} The average end-to-end inference time, including comprehensive Chain-of-Thought reasoning traces, is approximately 60s per SKU. By utilizing a batch size of 32 across the 16-GPU cluster, the system can process the entire catalog of 200,000 SKUs in approximately 7 hours.
    \item \textbf{Efficiency Comparison.} Under the same hardware constraints, the 235B model would require nearly 6$\times$ the computing time to achieve equivalent throughput, far exceeding the allocated 7-hour production window and hindering the feasibility of daily T+1 updates.
\end{itemize}

\textbf{Task-Specific Superiority.} 
Our results confirm that the 30B model, after SFT and DPO alignment, outperforms the 235B base model in pricing accuracy (EAMA 82.51\% vs. 67.07\%) and business value alignment (Q-Score 2.836 vs. 2.556). This confirms that for specialized e-commerce dynamic pricing task, a task-aligned compact model is more effective and resource-efficient than a massive general-purpose base model.




\end{document}